\documentclass[conference]{IEEEtran}
\IEEEoverridecommandlockouts  %
\usepackage{times}
\usepackage[numbers,sort&compress]{natbib}
\usepackage{xcolor,booktabs,tabularx,array,amsmath,graphicx,capt-of,placeins}
\usepackage[bookmarks=true,pdfauthor={Simon Won-Jin Kim, Hyung-Soon Park},pdftitle={Grasping in the Dark with In-Hand Proximity: Integrating and Verifying a New Modality in VLAs}]{hyperref}
\newcommand{\F}[1]{#1}
\makeatletter
\let\VPLA@maketitle\@maketitle
\renewcommand{\@maketitle}{\VPLA@maketitle
\begin{center}
\includegraphics[width=\textwidth]{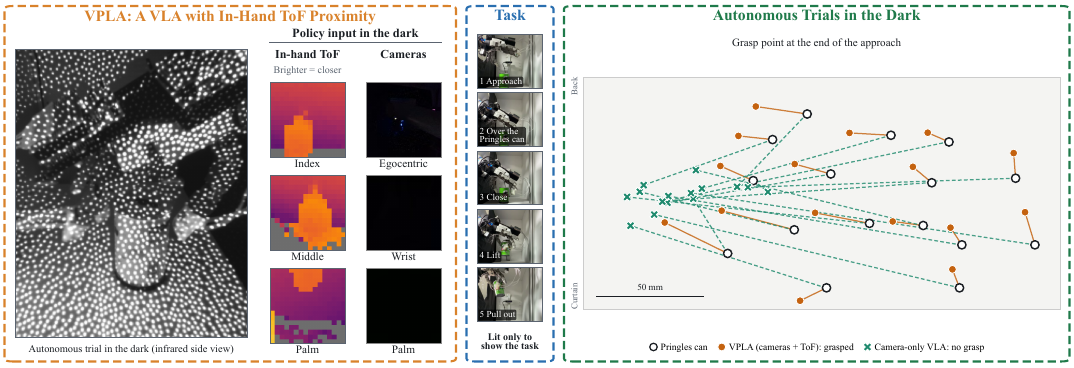}
\end{center}
\vspace{-0.8em}
\setcounter{figure}{0}%
\captionof{figure}{\textbf{The VPLA policy, a vision-language-action (VLA) model with in-hand time-of-flight (ToF) sensors.}
Left: the VPLA policy grasping the Pringles can in complete darkness (infrared view from a shelf depth camera with its projector
dots, not a policy input), with the ToF grids (brighter means closer) and nearly black camera images.
Middle: the task, lit for illustration (Fig.~\ref{fig:timestrip}).
Right: top view of the grasp points at the end of the approach for 16 Pringles can positions not seen in the demonstrations
(circles, Section~\ref{sec:results}).}
\label{fig:overview}
\vspace{1em}}
\makeatother
\begin{document}
\title{Grasping in the Dark with In-Hand Proximity:\\ Integrating and Verifying a New Modality in VLAs}
\author{\IEEEauthorblockN{Simon Won-Jin Kim$^{1}$ and Hyung-Soon Park$^{1}$%
\thanks{$^{1}$Neuro-Rehabilitation Engineering Laboratory, Department of Mechanical Engineering, KAIST, Daejeon 34141,
Republic of Korea {\tt\small \{simonkim02, hyungspark\}@kaist.ac.kr}}}}
\maketitle
\begin{abstract}
Vision-language-action (VLA) models rely on cameras, so they need another sensing modality when the camera images carry no
useful information.
For example, a camera-only VLA cannot locate an object in complete darkness.
In-hand time-of-flight (ToF) proximity sensors, which measure distance by timing how long infrared light takes to bounce off the target
object and return, could fill this gap.
To our knowledge, however, no VLA has yet used in-hand ToF proximity sensing together with vision.
We integrate ToF proximity sensing into the $\pi_{0.5}$ VLA model and call the result the vision-proximity-language-action (VPLA) policy.
We test the VPLA policy on taking a Pringles can out of a curtained shelf in complete darkness (0.0~lux), where the camera images carry
no information about the Pringles can's location.
In an offline exchange test, which replaces only the ToF input with the reading of another demonstration, the VPLA policy ignored
the ToF input under standard fine-tuning with the robot's joint angles in its prompt and a ToF encoder trained from scratch.
Removing the joint angles from the prompt, pre-training and freezing the ToF encoder, and collecting more demonstrations at the
edges and corners of the workspace together made the VPLA policy rely on the ToF input.
In blinded robot trials at 16 Pringles can positions not seen in the demonstrations, the VPLA policy approached, grasped and
pulled out the Pringles can in all 16 trials.
Two controls, a policy whose ToF readings were all marked invalid and a camera-only VLA, approached in 1 and 3 trials
(3 expected by chance) and never grasped.
In 16 further trials on the following night, in which the operator knew which policy was running, the VPLA policy under standard
fine-tuning approached in none.
\end{abstract}
\IEEEpeerreviewmaketitle
\section{Introduction}

\F{Vision-language-action (VLA) models rely on what their cameras see.}
On real robots, their success rate therefore falls as the scene gets darker~\cite{zhai2026evla,liu2026eventvla} and reaches zero well before the room is
completely dark~\cite{zhai2026evla}.
Sensors added to VLAs for low light also fall short in complete darkness.
For example, event cameras still need some light and a moving camera or object (Section~\ref{sec:related}).
\F{Thus, a robot that must grasp an object in complete darkness needs a sensor that works without room light and senses the
object before contact.}
Multizone time-of-flight (ToF) proximity sensors meet both requirements.
Each sensor emits pulses of infrared laser light, and an array of single-photon detectors times how long the light takes to return
from nearby surfaces.
Each sensor thus gives a grid of distances, one per zone.
Because the sensors fit in the palm or a fingertip of a robot hand, they can measure the hand-to-object distance before
contact~\cite{amsosram_tmf8829}.

To our knowledge, however, no VLA has used in-hand ToF proximity as an input.
\F{Prior work has added touch, force and other sensor modalities to VLAs and other robot policies~\cite{jones2025fuse,yu2025forcevla,zhang2026restacvla,niu2026trex,guo2025omnivla,motoda2026propra}.}
Most of these works add touch or force sensing, which responds only on contact, whereas ToF sensors can detect the object during
the approach.
However, adding another sensing modality to a policy does not ensure that the policy relies on it.
Fine-tuned with only the standard imitation loss, policies with added sensors often end up ``ignoring the new
modalities''~\cite{jones2025fuse}.
These works therefore add training losses or change how the sensor data enters the model (Section~\ref{sec:related}).
Yet they evaluate the resulting policies mainly by task success, which does not test whether the actions of a trained policy change
when only its sensor input changes.
As a result, it \F{remains an open problem} how to make a VLA rely on in-hand ToF proximity and how to verify that the VLA does.

\F{To address this problem, we present a vision-proximity-language-action (VPLA) policy, which integrates in-hand ToF proximity into the $\pi_{0.5}$ VLA model~\cite{intelligence2025pi05} through a separate transformer module.}
Our goal is a VLA that keeps its camera and language inputs and gains a new sensing modality that it actually uses.
To test whether the VPLA policy uses this new modality, we evaluate it on a task in which only the ToF sensors can locate the
object, namely taking a Pringles can out of a curtained shelf in complete darkness.

Making the policy rely on the ToF readings raises two challenges.
First, task success does not show which input a policy relies on, so we use an offline \emph{exchange test}, which measures
how far the planned grasp point moves when only a different ToF input is used.
Second, we found that the VPLA policy ignored the ToF input under \emph{standard fine-tuning}, in which the robot's joint angles are
written into the prompt and the ToF encoder is trained from scratch with the policy.
The ToF encoder is the network that turns the ToF readings into tokens.
In the exchange test, the planned grasp point of this policy did not move when a different ToF input was used.
We hypothesized that the joint angles, which are unique to each demonstration, allow the policy to identify each demonstration and
replay its recorded motion without reading the ToF input.
This proprioceptive shortcut is a form of causal confusion in imitation learning~\cite{dehaan2019causal,zhao2025statefree}.
\F{Removing the joint angles from the prompt, pre-training and freezing the ToF encoder, and collecting more demonstrations at the
edges and corners of the workspace together made the VPLA policy use the ToF readings, both in the exchange test and in autonomous
robot trials.}

\F{On the robot, we compare the VPLA policy with two controls trained on the same dataset, a policy whose ToF readings are all
marked invalid (the invalid-ToF control) and a camera-only VLA (Section~\ref{sec:setup}).}
In 48 blinded robot trials at 16 Pringles can positions not seen in the demonstrations, the VPLA policy approached, grasped and pulled
out the Pringles can in all 16 of its trials.
In contrast, the invalid-ToF control approached in one trial and the camera-only VLA in three (three expected by chance,
Section~\ref{sec:setup}).
Neither grasped the Pringles can in any trial.
In another round of 16 autonomous robot trials, the VPLA policy under standard fine-tuning approached in none.

In summary, our main contributions are
(i) the VPLA policy, to our knowledge the first VLA that takes in-hand ToF proximity as an input,
(ii) evidence that the known proprioceptive shortcut can cause a VLA to ignore a new sensing modality, and training
changes that prevented this in our task,
(iii) an exchange test that, unlike earlier input-permutation tests, replaces the ToF input with the real ToF reading of another training
demonstration to measure the effect of the added sensor offline, and
(iv) preliminary evidence from blinded robot trials that the policy uses the ToF information for our task in complete darkness.

\section{Related Work}
\label{sec:related}
\begin{figure*}[t]
\centering
\includegraphics[width=\textwidth]{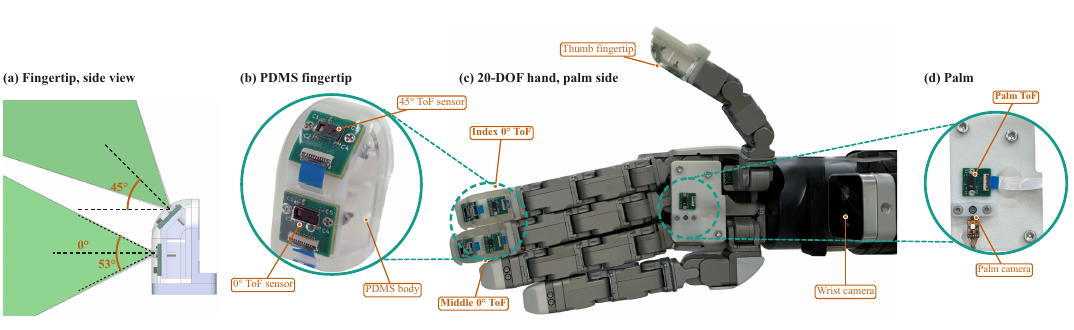}
\caption{\textbf{The hand and its ToF sensors.} (a) Fingertip side view (CAD model) with its 0$^\circ$ and 45$^\circ$ ToF sensors.
(b) A ToF fingertip in its silicone (PDMS) body.
(c) Palm side of the hand.
(d) Palm ToF sensor and palm camera.
Bold labels mark the three ToF sensors the VPLA policy uses. Cables and tape removed digitally.}
\label{fig:hardware}
\end{figure*}

\textbf{VLAs in low light.}
E-VLA~\cite{zhai2026evla} and Event-VLA~\cite{liu2026eventvla} add to a VLA an event camera, which reports changes in brightness and
therefore keeps working in far dimmer light than a standard camera.
In normal light, both works find that the VLA with the added event camera succeeds about as often as camera-only VLAs.
However, an event camera still needs some light and a change in the scene, such as a moving camera or object.
For example, E-VLA was tested only down to a few lux, where its success rate drops as sensor noise dominates.
Event-VLA reports failures when the object shows weak contrast or little motion.
Safe-Night VLA~\cite{yu2026safenight} instead adds a depth camera and a thermal camera, which needs a temperature contrast.
Depth cameras that project their own infrared pattern, such as the RealSense D400 and Orbbec Gemini 330 series, work in the dark
but measure only beyond a minimum distance from the camera.
Close-range models such as the RealSense D405 instead have no projector and need room light~\cite{intel_d400,orbbec_gemini330}.
Overall, none of these works shows a sensor locating a room-temperature object near the hand in complete darkness (Table~\ref{tab:related} in the Appendix).

\textbf{Added sensors that policies under-use.}
Simply adding a sensor as an extra policy input can hurt as well as help.
PROPRA~\cite{motoda2026propra} finds this for fingertip proximity and force sensing, and T-Rex~\cite{niu2026trex} finds that touch
signals fed directly to a pre-trained VLA can degrade its performance.
Similarly, FACTR~\cite{liu2025factr} finds that a policy given arm joint-torque readings in addition to camera images under-uses the
torque readings unless the images are blurred early in training.
Prior methods counter such under-use with training losses that tie the added sensor to language~\cite{jones2025fuse}
or to the robot state and next actions~\cite{motoda2026propra}.
FACTR instead makes this blur a curriculum that weakens as training proceeds.
Others change how the sensor enters the model, by using a mixture of experts~\cite{yu2025forcevla} or by feeding only the part of
the touch signal that vision fails to predict~\cite{zhang2026restacvla}.
OmniVLA~\cite{guo2025omnivla} and TAP-VLA~\cite{vandermerwe2026tapvla} draw the sensor signal onto the camera image, so that the
input format of the pre-trained VLA does not change.
T-Rex adds a separate tactile expert that completes the action denoising from touch.
None of these works uses ToF sensing.

\textbf{Proprioceptive shortcuts.}
Policies given proprioception as an input can fit their demonstrations through it instead of through the scene~\cite{lu2026visionproprio},
a form of causal confusion~\cite{dehaan2019causal}.
Further, removing this state input has been shown to improve the spatial generalization of such policies~\cite{zhao2025statefree}.
FingerEye~\cite{xu2026fingereye} reports that a policy can bypass fingertip sensing in the same way.
Under standard fine-tuning, the VPLA policy also used this proprioceptive shortcut, and we measure its effect directly.
Gat et al.~\cite{gat2021perceptual} measure a model's reliance on one input by permuting that input across test samples.
The exchange test instead replaces the ToF input with a real reading from another demonstration and measures the shift of the
planned grasp point (Section~\ref{sec:offline}).
The exchanged input thus stays within the training data.

\textbf{In-hand sensing before contact.}
Sensors in the fingers or palm perceive an object at close range before contact.
VLAs have taken depth from LiDAR and depth cameras that view the scene from a distance~\cite{li2025pointvla,chen2026pointact}, but none
from sensors in the hand.
Lancaster et al.~\cite{lancaster2018improved} and Chang et al.~\cite{chang2026toftouch} place single-zone and multizone ToF sensors
in the fingers of parallel grippers but use hand-designed controllers.
Heo and Park~\cite{heo2024tmech} use in-hand ToF sensing to choose the grasp posture of a dexterous prosthetic hand, which the
user opens and closes through electromyography (EMG).
TetraGrip~\cite{torrado2025tetragrip} feeds the readings of multizone ToF sensors on a suction gripper to a reinforcement-learning policy without
camera input.
PROPRA~\cite{motoda2026propra} feeds optical distance and force readings from the fingertips of a parallel gripper,
with wrist and scene camera images, to a diffusion policy.

FingerViP~\cite{zhang2026fingervip} and FingerEye~\cite{xu2026fingereye} instead place RGB cameras in the fingertips of
multi-fingered hands.
These cameras have no light of their own, so FingerViP names built-in illumination as future work, and FingerEye avoids such illumination by design.
Overall, none of these works feeds its in-hand sensor readings into a VLA or tests proximity sensing in complete darkness.
The VPLA policy addresses both gaps.
It is a VLA that reads multizone ToF sensors in the palm and two fingertips of a five-finger hand, and we test it in complete
darkness (Section~\ref{sec:experiments}).

\section{System}
\label{sec:system}

\begin{figure*}[t]
\centering
\includegraphics[width=\textwidth]{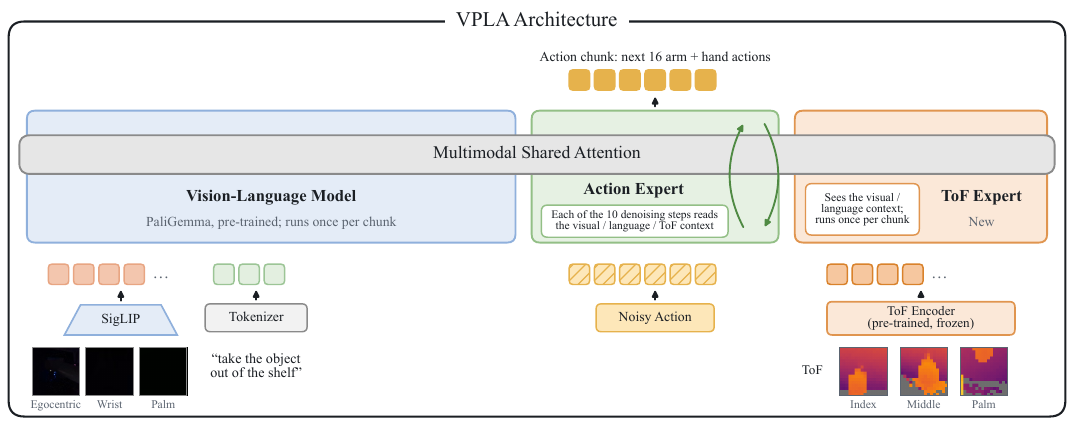}
\caption{\textbf{VPLA policy architecture.} $\pi_{0.5}$ plus the ToF expert, which reads the tokens of the pre-trained, frozen ToF encoder.
The three experts share the self-attention layers. The action expert reads the visual, language and ToF context at each of its
10 denoising steps (circular arrows). The inputs shown are from one robot trial in complete darkness.}
\label{fig:arch}
\end{figure*}

\subsection{Time-of-Flight Fingertips and Palm}
\label{sec:hardware}
The robot is a six-degree-of-freedom (DOF) arm (Fairino FR10) with a 20-DOF five-finger hand (Tesollo DG-5F).
The policy receives $224\times224$ RGB images from three cameras, an egocentric camera (RealSense D455), a wrist camera (RealSense D405) and a miniature USB camera in the palm.
Seven ToF sensors (TMF8829) sit in the hand, two in each of our custom thumb, index and middle fingertips (Fig.~\ref{fig:hardware}a)
and one in the palm.
Each sensor divides its field of view into a $16\times16$ grid of \emph{zones} and measures one distance per zone (settings in Appendix~\ref{app:tof}).
A zone's reading is marked invalid when its reflected signal is too weak, for example when no surface lies within range.
Readings closer than 8~mm, below the sensor's specified minimum of 10~mm, are also marked invalid.
The policy reads three of the sensors, the palm sensor and the palm-facing (0$^\circ$) sensors of the index and middle fingertips, because only these three face the top
of the Pringles can at the start of each episode.

\subsection{Scene and Demonstrations in the Dark}
\label{sec:demos}
A Pringles can stands on a felt panel in a curtained shelf.
A motorized stage under the panel moves the Pringles can sideways ($x$) and in depth ($y$, away from the robot), and the demonstrations cover
an area of 200~mm in $x$ by 100~mm in $y$.
With the room lights off, a lux meter in the shelf beside the hand read 0.0~lux with the LEDs of the hand and the ToF boards
uncovered, as in the demonstrations and trials.
In this darkness, the Pringles can is not visible in the camera images (Appendix~\ref{app:dark}).
Every episode starts from the same hand pose inside the shelf, which we call the \emph{start pose}.

A human operator recorded the demonstrations by teleoperation in the dark, guided by two depth cameras in the shelf and a drawn
top-view guide, which help only the operator and are never inputs to the policy (Appendix~\ref{app:teleop}).
The infrared projectors of these depth cameras, whose dots appear in Fig.~\ref{fig:overview}, stayed on in every demonstration
and in the trials of every policy.

We recorded 150 demonstrations in three rounds and trained on 144 of them (113{,}177 frames at 30~Hz).
The first round placed the Pringles can across the workspace (54 of 60 used, Appendix~\ref{app:teleop}).
The second added 40, mostly at the edges and corners of the workspace, where a pilot policy trained on the first 54 demonstrations had missed the Pringles can.
The third added 50 at the positions farthest from all earlier demonstrations, mostly at the four corners and the two side edges.

\subsection{The VPLA Policy}
\label{sec:policy}
\textbf{Observations and actions.}
The policy receives the three camera images, the fixed prompt ``take the object out of the shelf'' and the distance grids of the three ToF sensors.
It outputs an \emph{action chunk}, the next 16 actions (0.53~s at 30~Hz).
The robot executes the first 8 actions before the policy plans again.
Each action consists of an arm end-effector pose relative to the chunk's starting arm pose and 20 absolute hand joint angles.
Inverse kinematics converts the arm pose to joint commands.

\textbf{Architecture.}
$\pi_{0.5}$ has a PaliGemma vision-language model~\cite{beyer2024paligemma}, which turns the images and the prompt into tokens
and uses the SigLIP image encoder~\cite{zhai2023siglip} for the images.
$\pi_{0.5}$ also has an action expert, which produces the action chunk by flow matching in 10 denoising steps.
The VPLA policy adds a third expert, the \emph{ToF expert} (Fig.~\ref{fig:arch}).
This expert is a randomly initialized transformer with the Gemma architecture~\cite{team2024gemma} (18 layers, width 512, 99\,M parameters).
The three experts interact only through shared self-attention layers, under a blockwise causal attention mask as in
$\pi_0$~\cite{black2024pi0}.
Thus, ToF tokens attend to the vision-language tokens and to each other, and action tokens attend to all tokens.

\textbf{ToF encoder.}
Before encoding, each distance $d$ is normalized to a \emph{closeness} $c=\mathrm{clip}(1-d/D,\,0,\,1)$ between 0 and 1, with $D=300$~mm.
A near surface gives a value close to 1, and invalid zones and surfaces 300~mm or farther away give 0.
We chose $D=300$~mm because the felt panel lies about 313~mm from the palm sensor at the start pose.
The panel's closeness is thus 0, and only objects on it, such as the Pringles can, give a nonzero closeness.
The encoder $E_\phi$, whose trainable weights are $\phi$, splits each $16\times16$ grid into 16 blocks of $4\times4$ zones and turns each block into one token.
A shared multilayer perceptron reads five features per block, the mean closeness of each of its four $2\times2$ sub-blocks and the
fraction of valid zones.
Learned embeddings mark each token's sensor and grid position, and the three sensors give 48 tokens.

\textbf{Training changes.}
The \emph{baseline VPLA policy} uses standard fine-tuning, with the joint angles in the prompt and a ToF encoder trained from scratch with the policy.
Two changes make the VPLA policy rely on the ToF readings.
(1) \emph{No joint angles in the prompt.} $\pi_{0.5}$ writes the robot's joint angles into the prompt as text, and we remove
them because they give the baseline VPLA policy a shortcut that lets it ignore the ToF readings (Section~\ref{sec:offline}).
(2) \emph{A pre-trained, frozen ToF encoder.} Before policy training, the encoder $E_\phi$ and a regression head $h$ are trained
alone to estimate the Pringles can's stage position $\mathbf{p}=(x,y)$ from the closeness grids $C$ of the three sensors at the
start pose.
The encoder and the head minimize the MSE over the $K$ training readings,
\begin{equation}
\min_{\phi,\,h}\ \frac{1}{K}\textstyle\sum_{k=1}^{K} \big\lVert h\big(E_\phi(C_k)\big)-\mathbf{p}_k\big\rVert^2 ,
\label{eq:pretrain}
\end{equation}
where $\lVert\cdot\rVert$ is the length of the position error in $x$ and $y$, and the labels $\mathbf{p}_k$ are the positions set
by the motorized stage.
For the encoder dataset, we collected 3{,}166 readings at 201 positions, spaced about 15~mm apart, and 57 readings in the dark
at 56 positions (Appendix~\ref{app:training}).
The head is then removed, and the encoder is frozen.
The encoder was not trained on the first 54 demonstrations of the teleoperation dataset.
On these demonstrations, its estimate of the Pringles can's stage position, averaged per demonstration, has an RMS
error of 4.69~mm in $x$ and 4.12~mm in $y$.

\section{Experiments}
\label{sec:experiments}
We test whether the VPLA policy uses the ToF readings to find the Pringles can in the dark, offline and in autonomous robot trials.

\subsection{Experimental Setup}
\label{sec:setup}
\label{sec:task}
\textbf{Policies.} The following four policies are fine-tuned in full from $\pi_{0.5}$ on the same 144 demonstrations for
24{,}350 steps (about 7 epochs, Appendix~\ref{app:training}).
Besides the \emph{VPLA policy} (Section~\ref{sec:policy}), the \emph{invalid-ToF control} is a policy trained and tested like the VPLA policy but
with every ToF sensor zone marked invalid. A fully invalid ToF input never occurs in the demonstrations, so it cannot simply be fed to the VPLA policy.
The \emph{camera-only VLA} is $\pi_{0.5}$ without ToF readings or joint angles in the prompt.
The \emph{baseline VPLA policy} uses standard fine-tuning (Section~\ref{sec:policy}).
In addition to these four, two \emph{proximity-language-action (PLA) policies} are trained like the VPLA policy but with all camera images masked, one on the
first 94 and one on all 144 demonstrations.
They can thus locate the Pringles can only through the ToF sensors.
We do not compare with event or infrared camera VLAs, because we ask whether a VLA can be made to use in-hand ToF sensors, not which
sensor works best in the dark.

\textbf{Task.} The policy must take the Pringles can out of the curtained shelf (Fig.~\ref{fig:setup}) in complete darkness
(Fig.~\ref{fig:timestrip} in the Appendix).
Each trial allows a single grasp attempt, so the policy must locate the Pringles can before touching it rather than searching the shelf by touch.
The Pringles can stands at 16 test positions not seen in the demonstrations, 11.7 to 22.1~mm from the nearest demonstration position
and within 8~mm of a position used to pre-train the encoder (Fig.~\ref{fig:overview}, right, and Fig.~\ref{fig:positions} in the
Appendix).
The VPLA policy and the two controls each run once per position, 48 trials in random order under code names.
The baseline VPLA policy runs at the same positions in the same order on the following night without code names.
Positions, order, outcome rule and tests were pre-registered before each night's trials (Appendix~\ref{app:protocol}).

\textbf{Outcomes.} The main outcome, the \emph{approach}, is computed from the arm's joint-angle log of each trial.
The \emph{grasp point} is the hand's tool center point (TCP), 216~mm from the arm's flange.
It lies over the Pringles can just before the fingers close in the demonstrations.
The approach ends when the grasp point has descended 50~mm, and it succeeds if the descent ends within 30~s of the trial start with
the grasp point within 40~mm of the Pringles can in $x$.
The two control policies come down at almost the same spot near the left wall of the shelf in every trial, regardless of the Pringles can's position.
Their approach therefore counts as a success whenever the Pringles can happens to stand within 40~mm of that spot, although the
control does not locate the Pringles can.
For the spot where the controls came down in an earlier session, this is the case at 3 of the 16 test positions, which we
pre-registered as the chance level.
Because the approach count alone cannot show that a policy located the Pringles can, we also report where the grasp point
lands at the end of the approach relative to the Pringles can, in $x$ and $y$.
The operator scores the \emph{grasp}, the Pringles can held after the fingers first close, and the \emph{pull-out}, the Pringles can taken out of the shelf.

\textbf{Hypotheses.} H1 is that \emph{under standard fine-tuning, the baseline VPLA policy ignores the ToF input even with
all 144 demonstrations}.
H2 is that \emph{the training changes of Section~\ref{sec:policy} make the policy use the ToF input}.
H3 is that \emph{the VPLA policy approaches the Pringles can at more positions than either control}.
\begin{figure}[t]
\centering
\includegraphics[width=\columnwidth]{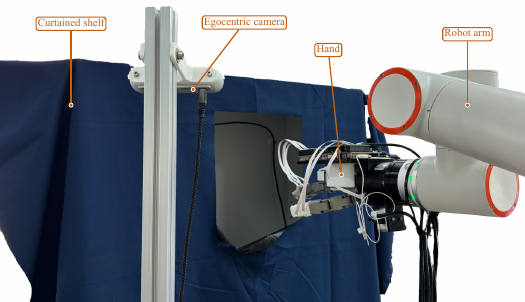}
\caption{\textbf{Experimental setup, lit for illustration} (all trials ran in complete darkness). The egocentric camera stands
outside the curtained shelf, and the robot arm reaches with the hand through an opening in the curtain. The background was removed
digitally for clarity.}
\label{fig:setup}
\end{figure}
\subsection{Offline Check: Does the Policy Use the ToF Input?}
\label{sec:offline}
\begin{table}[t]\footnotesize
\centering
\caption{\textbf{Without joint angles in the prompt, the use of the ToF input grows with each training change.} ToF sensitivity in the exchange test
(\%~of the Pringles can displacement, $x$ / $y$). Intervals and more rows are in Table~\ref{tab:ablation_full}.}
\label{tab:ablation}
\setlength{\tabcolsep}{4pt}
\begin{tabular}{@{}l l c c c@{}}
\toprule
Joint angles & Encoder & Demos & ToF sensitivity (\%) & Robot \\
\midrule
in prompt & from scratch & 54 & 0.0 / 0.0 & -- \\
in prompt & frozen & 54 & 0.0 / 0.0$^a$ & -- \\
none & from scratch & 54 & 1.5 / 1.3 & -- \\
none & frozen & 54 & 3.3 / 3.7 & 9/16 \\
\midrule
none & frozen & 94 & 6.7 / 8.4 & 13/16 \\
in prompt & from scratch & 144 & 0.0 / 0.0 & 0/16 \\
in prompt & frozen & 144 & 0.1 / 0.2 & -- \\
none & from scratch & 144 & 6.5 / 5.1 & 7/16 \\
none & pre-trained & 144 & 7.6 / 9.0 & -- \\
none & frozen & 144 & \textbf{8.7 / 9.0} & \textbf{16/16} \\
\midrule
\multicolumn{3}{@{}l}{Human demonstrations} & 11.1 / 14.9 & -- \\
\bottomrule
\end{tabular}
\\[2pt]
\raggedright\scriptsize
An encoder trained from scratch starts from random weights and learns only together with the policy. A frozen encoder is pre-trained
(Section~\ref{sec:policy}) and then kept fixed, and a pre-trained encoder is not frozen but trained further with the policy. Training ran for 10 epochs with 54 and 94 demonstrations and for about
7 with 144 (Appendix~\ref{app:training}). The Robot column gives the successful approaches at the 16 test positions of Section~\ref{sec:setup}, from three
nights (Section~\ref{sec:results}).
$^a$Training starts each demonstration 1.5~s instead of 0.5~s before the hand moves (15{,}290 steps).
\end{table}
\label{sec:offline_results}
The \emph{exchange test}, a counterfactual input test, keeps a training frame's camera images and prompt and replaces only its ToF input with the ToF reading of another
demonstration whose Pringles can stood at a different position.
The test uses 52 of the 54 demonstrations of the first round (Section~\ref{sec:demos}).
The other two are left out because their hand never moved faster than 20~mm/s horizontally before it closed on the Pringles can.
Thus, the start of their motion, from which the test frames are timed, is not defined (Appendix~\ref{app:exchange}).
From each demonstration we take three test frames, at 0.5~s and 0.27~s before the hand starts to move and at the moment it starts.
At each test frame, the policy first plans its action chunk with the frame's own ToF input, which gives the \emph{reference plan}.
It then plans again with the ToF input of each of 9 other demonstrations whose Pringles can stood at a different position, which gives 9
\emph{exchanged plans}.
Each exchanged ToF input is taken at the same moment relative to the start of that demonstration's hand motion.
For exchange $i$, let $\Delta_i$ be how far the planned grasp point at the last step of the action chunk moves from the reference
plan to the exchanged plan, and $d_i$ the displacement of the Pringles can between the two demonstrations, along $x$ or $y$.
The \emph{ToF sensitivity} is the least-squares slope through the origin,
\begin{equation}
s=\textstyle\sum_i d_i\,\Delta_i \big/ \sum_i d_i^2 ,
\label{eq:sensitivity}
\end{equation}
which we give in percent.
For example, a sensitivity of 8.7\,\% in $x$ means that when the exchanged ToF input comes from a demonstration whose Pringles can
stood 100~mm farther in $x$, the planned grasp point moves 8.7~mm in the same direction within one action chunk.
A policy that ignores the ToF input gives 0, and the recorded motion of the human demonstrations over the same frames gives
11.1\,\% in $x$ and 14.9\,\% in $y$ (Table~\ref{tab:ablation}).
Because $\pi_{0.5}$ generates each action chunk from random noise, each plan averages the chunks from 4 noise samples.
The reference plan and every exchanged plan start from the same 4 samples, so that $\Delta_i$ comes only from the ToF input.
The 95\,\% intervals come from 2{,}000 bootstrap resamples, each drawing 52 demonstrations with replacement (Appendix~\ref{app:exchange}).

\textbf{The baseline VPLA policy ignores the ToF input (H1).}
The ToF sensitivity of the baseline VPLA policy is 0.0\,\% in $x$ and in $y$ (95\,\% interval within 0.01\,\% of zero).
This is not because the ToF readings lack the information, since a linear regression on the raw ToF readings predicts the Pringles
can's offset from the grasp point on held-out demonstrations with an $R^2$ of 0.94 in $x$ and 0.96 in $y$.
These values leave out one outlier demonstration, in which all three ToF sensors read a surface about 110~mm away.
With this demonstration included, $R^2$ is 0.91 in $x$ and 0.70 in $y$.
Instead, the joint angles in the prompt offer a shortcut.
The joint-angle text in the prompt is unique to each of the 52 tested demonstrations.
In a single check, exchanging only this text for the joint-angle text of another demonstration moved the planned grasp point by up
to 15.5~mm along one axis.

Thus, the baseline VPLA policy likely recognizes each demonstration by its joint angles and imitates that demonstration's recorded
motion without using the ToF readings, a behavior known as a proprioceptive shortcut.
Heo and Park~\cite{heo2024tmech} likewise leave joint angles out of a grasp posture decoder because joint angles can bias the decoder to follow
its previous decision.
Similarly, Grant et al.~\cite{grant2026features} find that a fine-tuned $\pi_{0.5}$ can execute a trajectory bound
to absolute workspace coordinates while ignoring the visible scene.
With the joint angles in the prompt and a pre-trained, frozen encoder, the ToF sensitivity stays near zero, at 0.0\,\% in $x$ and $y$ with 54
demonstrations and at 0.1 and 0.2\,\% with 144 (Table~\ref{tab:ablation}).

\textbf{The training changes make the policy use the ToF input (H2).}
We give the effects of the training changes in the order in which we made them (Table~\ref{tab:ablation}).
First, removing the joint angles from the prompt already gives a nonzero sensitivity, 1.5\,\% in $x$ and 1.3\,\% in $y$ with 54
demonstrations, even though the encoder is still trained from scratch with the policy.
Second, a pre-trained, frozen encoder more than doubles the sensitivity with the same 54 demonstrations, to 3.3 and 3.7\,\%.
Third, more demonstrations raise it further, to 6.7 and 8.4\,\% with 94 and to 8.7\,\% (95\,\% interval 7.7 to 9.7) and
9.0\,\% (7.5 to 10.6) with all 144.
With 144 demonstrations, we also separated the effect of pre-training the encoder from that of freezing it.
An encoder trained from scratch gives 6.5\,\% (5.5 to 7.5) and 5.1\,\% (4.1 to 6.2), and a pre-trained encoder that is not frozen but
trained further with the policy gives 7.6\,\% (6.8 to 8.5) and 9.0\,\% (7.7 to 10.4).
In $y$, pre-training thus accounts for the whole gain from 5.1 to 9.0\,\%, and freezing adds nothing.
In $x$, pre-training and freezing each add about 1.1 percentage points, which is less than the width of the 95\,\% intervals.
Overall, removing the joint angles enabled the use of the ToF input in the exchange test, and the pre-trained encoder and more demonstrations
strengthened this use.

\begin{figure*}[t]
\centering
\includegraphics[width=\textwidth]{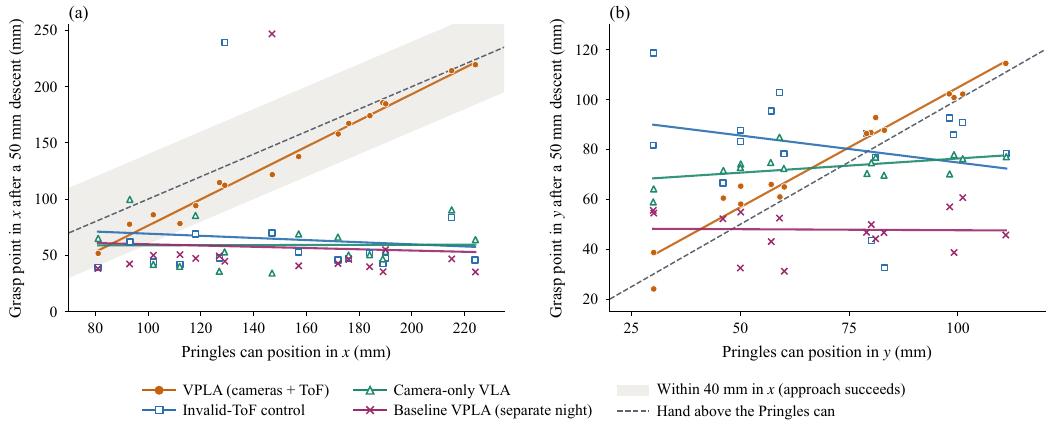}
\caption{\textbf{Does the hand follow the Pringles can?} Grasp-point position at the end of the approach (after a 50~mm descent)
against Pringles can position, in $x$ (a) and in $y$ (b), with one marker per trial and least-squares fit lines. All trials were
blinded except those of the baseline VPLA policy, which ran on a separate night without code names.}
\label{fig:deviation}
\end{figure*}
\subsection{Robot Results in Complete Darkness}
\label{sec:results}
\textbf{Success rates.} The VPLA policy approached, grasped and pulled out the Pringles can at all 16 positions (95\,\%
Clopper-Pearson interval 79 to 100\,\%), whereas the invalid-ToF control and the camera-only VLA never grasped it
(Table~\ref{tab:robot}).
The two controls approached at 1 and 3 positions, the chance level, and both differences in approach are significant after Holm
correction (H3).
Because the VPLA policy and the invalid-ToF control differ only in the ToF information, the difference in their results can be attributed to the ToF input.
No trial was discarded or stopped by the operator (protocol and safety limits in Appendix~\ref{app:protocol}).
The baseline VPLA policy, trained on the same 144 demonstrations, approached the Pringles can at none of the 16 positions
($p = 3.1\times10^{-5}$ against the VPLA policy, Table~\ref{tab:robot}), which supports H1 on the robot and agrees with the exchange test.
Because the two policies differ only in the removed joint angles and the pre-trained, frozen encoder, the VPLA policy's use of
the ToF input can be attributed to the combination of these changes rather than to more demonstrations (H2).

\begin{table}[t]\footnotesize
\centering
\caption{\textbf{Robot trials in complete darkness} (16 positions, one trial each). $p$ is from the exact McNemar test of the
approach against the VPLA policy, paired by position.}
\label{tab:robot}
\begin{tabular}{@{}l c c c c@{}}
\toprule
Policy & Approach & Grasp & Pull-out & $p$ \\
\midrule
VPLA policy & \textbf{16/16} & \textbf{16/16} & \textbf{16/16} & -- \\
Invalid-ToF control & 1/16 & 0/16 & 0/16 & $6.1\times10^{-5}$ \\
Camera-only VLA & 3/16 & 0/16 & 0/16 & $2.4\times10^{-4}$ \\
Baseline VPLA policy$^a$ & 0/16 & 0/16 & 0/16 & $3.1\times10^{-5}$ \\
\bottomrule
\end{tabular}
\\[2pt]
\raggedright\scriptsize
$^a$Following night, same positions and order, unblinded.
\end{table}

\textbf{Does the hand follow the Pringles can?}
A policy that locates the Pringles can ends its approach with the grasp point over the Pringles can.
Across the 16 test positions, its grasp point thus moves with the Pringles can at a slope of 1, shown as the dashed line in
Fig.~\ref{fig:deviation}.
A policy whose hand comes down at one spot regardless of the Pringles can's position instead gives a slope near 0.
In $x$ (Fig.~\ref{fig:deviation}a), the VPLA policy's grasp point ended a median 15~mm from the Pringles can, with a least-squares
slope of 1.17 (95\,\% interval 1.10 to 1.25).
In contrast, the grasp points of the invalid-ToF control, the camera-only VLA and the baseline VPLA policy ended a median 107, 98 and 108~mm from the
Pringles can, with slopes of $-0.09$, 0.00 and $-0.06$.
In $y$ (Fig.~\ref{fig:deviation}b), the VPLA policy's slope was 0.96 (0.83 to 1.10), against $-0.22$, 0.11 and $-0.01$ for the other three policies.
Its grasp point ended a median 6~mm behind the Pringles can, close to the 9~mm of the human demonstrations at the same moment.
The few successful approaches of the two controls all came at the leftmost positions (Pringles can at $x$ = 81 to 118~mm), as expected by chance
(Section~\ref{sec:setup}).

\textbf{Which training changes matter on the robot?}
The exchange test showed that the use of the ToF input grows with each training change (H2).
To test whether this also holds on the robot, we ran three more policies blinded under code names at the same 16 positions on a
later night (Table~\ref{tab:ablation}).
Like the VPLA policy, all three have no joint angles in the prompt, and each differs from the VPLA policy in just one training change.
Two keep the pre-trained, frozen encoder but use only the first 94 or the first 54 demonstrations.
The third keeps all 144 demonstrations but uses an encoder trained from scratch.
Each policy's approaches are compared with those of the VPLA policy by the exact McNemar test with Holm correction.
\emph{Fewer demonstrations.} The 94-demonstration policy approached and grasped at 13 positions and pulled out the Pringles can
at 11.
Its 13 approaches are not significantly fewer than the 16 of the VPLA policy ($p = 0.25$).
The 54-demonstration policy approached at only 9 positions ($p = 0.016$, significant) and grasped at 2.
Likewise, the grasp point of the 94-demonstration policy followed the Pringles can in $x$ at a slope of 0.75 ($r = 0.79$),
whereas the 54-demonstration policy's slope of 0.57 was not significant.
In the exchange test, the ToF sensitivity in $x$ fell in the same order, from 8.7\,\% with 144 demonstrations to 6.7\,\% with 94
and 3.3\,\% with 54.
\emph{Encoder trained from scratch.} With all 144 demonstrations, the policy whose encoder was trained from scratch approached,
grasped and pulled out the Pringles can at only 7 positions ($p = 0.0039$, significant).
Its grasp point still moved with the Pringles can in $x$ at a slope of 0.94 ($r = 0.67$).
However, in 7 of its 9 failed approaches it came down 57 to 114~mm to the left of the Pringles can, and in the other 2 it never
descended 50~mm.
Thus, with the same 144 demonstrations, the pre-trained, frozen encoder raised the approaches from 7 to 16 positions, although
this comparison spans two nights with one training seed per policy.
\emph{Exchange test and robot.} Across the five policies tested on the robot, the approach rate rose with the ToF sensitivity in
$x$, but not significantly (Spearman $\rho = 0.90$, $p = 0.083$).
The one exception is the policy whose encoder was trained from scratch, which had a higher sensitivity than the
54-demonstration policy (6.5 against 3.3\,\%) but approached at fewer positions (7 against 9).

\textbf{Without camera images.} In unblinded trials, the PLA policies trained on 94 and on 144 demonstrations both approached the
Pringles can at all 16 positions and grasped and lifted it in 12 and 14 trials, respectively.
A safety limit on the hand's height ruled out the pull-out in these trials (Appendix~\ref{app:protocol}).

\textbf{Beyond the demonstrated area.} We also tested whether the policies found the Pringles can outside the area covered by the
demonstrations.
At 8 positions 46 to 51~mm to the right of the rightmost demonstrated Pringles can position (Fig.~\ref{fig:positions} in the
Appendix), the VPLA policy and the two PLA policies each ran once in unblinded trials, and none of them approached the Pringles can
in these 24 trials.
In 19 of these trials the hand moved toward the left wall, away from the Pringles can, and in 4 the VPLA policy's hand came down 18 to
39~mm short of the Pringles can in $x$.

\section{Conclusion}
\label{sec:conclusion}
We integrated in-hand ToF proximity sensors into a VLA model and tested whether the resulting VPLA policy uses the ToF input in a
grasping task in complete darkness, where the cameras cannot see the Pringles can.
An offline exchange test showed that under standard fine-tuning the policy
ignored the ToF input, likely through a proprioceptive shortcut based on the joint angles in its prompt.
Removing the joint angles, using a pre-trained, frozen ToF encoder and collecting more demonstrations made the policy use the ToF input in the
exchange test and in blinded robot trials.
To find which change had the largest effect, we trained ten policies with different combinations of these three changes and tested
all ten with the exchange test and five on the robot.
Removing the joint angles enabled the use of the ToF input, and the pre-trained, frozen encoder and the added demonstrations
strengthened this use.
Across the five policies tested on the robot, the approach rate rose with the ToF sensitivity, although not significantly.

\section{Limitations and Future Work}
\label{sec:limits}
(1) The experiments cover one object, task, prompt and training seed, with 16 trials per policy and none in normal light, so the task exercises
neither the language nor the vision of the VLA.
(2) The operator who scored the grasp and pull-out could tell the VPLA policy from the controls by how each policy approached, so these two scores
may not be fully blind (Appendix~\ref{app:protocol}).
(3) The 16 test positions lie 11.7 to 22.1~mm from the nearest demonstration position, and we chose them after adding 90 demonstrations,
mostly at the workspace edges and corners where a pilot policy had failed.
Replaying the demonstration whose start-pose ToF reading is closest to the reading at the test position would also succeed in the
approach at 15 or 16 of the 16 test positions.
Thus, the trials at these positions show that the policy reads the ToF input in the dark but not that it generalizes beyond its training demonstrations.
In line with this, neither the VPLA policy nor the two PLA policies approached the Pringles can at the 8 positions beyond the
demonstrated area
(Section~\ref{sec:results}).
(4) The ToF encoder was pre-trained and tested only with the hand at the start pose, and its pre-training needs Pringles can
positions as labels, which every new object would also need.
(5) Only five of the ten policies in Table~\ref{tab:ablation} ran on the robot, and the others, including the one whose
pre-trained encoder is not frozen, were tested only offline.
Across these five policies, the approach rate did not correlate significantly with the ToF sensitivity, and the additional
demonstrations also lengthened training.
(6) The lux meter read 0.0~lux, but uncompressed wrist and palm frames (which are not inputs to the policy) show the Pringles can very
faintly (Appendix~\ref{app:dark}).
(7) Like other optical proximity sensors~\cite{lancaster2018improved}, the ToF sensors can fail to detect transparent or highly
specular objects, and we tested no such objects.

\textbf{Future work.} We will train on several objects, each named in its own prompt, and test whether the policy approaches the
object that the prompt specifies when multiple objects are on the panel.
We will also test the VPLA policy at Pringles can positions outside the demonstrated area and on objects that do not appear in
its demonstrations.
To test positions between demonstrations, we have trained a VPLA policy on 88 of the 144 demonstrations, chosen so that two regions
inside the workspace hold no demonstrations.
We will test this policy at Pringles can positions 35 to 43~mm from the nearest of these 88 demonstrations.
Finally, we will run robot trials of the policy whose pre-trained encoder is not frozen, and add training seeds and trials.

\FloatBarrier %
\bibliographystyle{unsrtnat-etal}
\bibliography{references}
\appendix
\section{Supplementary Material}
\subsection{Comparison with the Closest Work}
\label{app:related}
Table~\ref{tab:related} compares the VPLA policy with the closest prior work discussed in Section~\ref{sec:related}.
For each work, it lists the added sensing, the policy, the darkest condition tested and what the work does so that the policy
relies on the added sensor.

\begin{table*}[t]
\centering
\caption{\textbf{Closest prior work.} Lowest light: the darkest condition in which the policy was tested. Method against under-use: what the work does so that the policy relies on the added
sensor. --: not applicable or not tested.}
\label{tab:related}
\footnotesize
\setlength{\tabcolsep}{4pt}
\begin{tabular}{@{}l>{\raggedright\arraybackslash}p{3.3cm}>{\raggedright\arraybackslash}p{3.0cm}>{\raggedright\arraybackslash}p{3.0cm}>{\raggedright\arraybackslash}p{4.4cm}@{}}
\toprule
Work & Added sensing & Policy & Lowest light & Method against under-use \\
\midrule
E-VLA~\cite{zhai2026evla} & event camera & VLA (SmolVLA-style~\cite{shukor2025smolvla}) & 2~lux & camera image randomly dropped in training (50\,\%) \\ \addlinespace
Event-VLA~\cite{liu2026eventvla} & event camera & VLA (OpenVLA-based~\cite{kim2024openvla}) & ``near-dark'', no lux given & auxiliary loss predicting future event maps \\ \addlinespace
Safe-Night VLA~\cite{yu2026safenight} & thermal camera, depth & VLA (GR00T N1.5, an update of N1~\cite{bjorck2025gr00t}) & RGB dimmed in software & augmentation of the RGB image only \\ \addlinespace
FuSe~\cite{jones2025fuse} & touch, audio & generalist policy (Octo~\cite{team2024octo}) & inside a paper bag (dim wrist camera), no lux given & contrastive and language-generation losses \\ \addlinespace
FACTR~\cite{liu2025factr} & arm joint torques & action chunking transformer & -- & camera images blurred early in training, less so as training proceeds \\ \addlinespace
ForceVLA~\cite{yu2025forcevla} & estimated wrench & VLA ($\pi_0$~\cite{black2024pi0}) & -- & force token fused after the vision-language model by a mixture-of-experts module \\ \addlinespace
T-Rex~\cite{niu2026trex} & fingertip touch (force, deformation) & VLA (Qwen3-VL backbone) & -- & separate tactile expert that completes the denoising, and a partly frozen, pre-trained encoder \\ \addlinespace
ResTacVLA~\cite{zhang2026restacvla} & touch (GelSight Mini) & VLA ($\pi_{0.5}$~\cite{intelligence2025pi05}) & -- & residual tactile representation, gated by visual uncertainty \\ \addlinespace
PROPRA~\cite{motoda2026propra} & reflective optical proximity, touch & diffusion policy & -- & contrastive encoder pre-training against robot state and next actions \\ \addlinespace
OmniVLA~\cite{guo2025omnivla} & thermal camera, mmWave radar, microphones & VLA (SmolVLA~\cite{shukor2025smolvla}) & -- & sensor maps painted onto the camera image inside object outlines segmented from that image \\ \addlinespace
TAP-VLA~\cite{vandermerwe2026tapvla} & two GelSight touch sensors on a parallel gripper & VLA ($\pi_{0.5}$~\cite{intelligence2025pi05}) & -- & tactile shear drawn as arrows onto the camera images \\ \addlinespace
FingerViP~\cite{zhang2026fingervip} & fingertip cameras & diffusion policy & non-illuminated button in a box open on one side, no lux given & -- \\ \addlinespace
TetraGrip~\cite{torrado2025tetragrip} & multizone ToF sensors on four suction cups & reinforcement-learning policy, no camera & -- & -- \\ \addlinespace
Chang et al.~\cite{chang2026toftouch} & multizone ToF sensing and optical touch in a parallel-gripper finger & hand-designed controllers & -- & -- \\
\midrule
\textbf{Ours (VPLA policy)} & palm and fingertip ToF grids & VLA ($\pi_{0.5}$~\cite{intelligence2025pi05}) & complete darkness, 0.0~lux & no joint angles in the prompt, pre-trained, frozen ToF encoder, demonstrations at workspace edges and corners \\
\bottomrule
\end{tabular}
\end{table*}

\subsection{ToF Sensor Settings}
\label{app:tof}
Two STM32 Nucleo-H503RB boards read the seven TMF8829 sensors, one board the four index and middle fingertip sensors and the other
the palm and two thumb sensors.
The sensors on each board share one I3C bus, and the board triggers them sequentially.
Each sensor runs the manufacturer's firmware (version 1.2.194) with the $16\times16$-zone high-accuracy preset.
Each sensor has an 80$^\circ$ diagonal field of view ($67.9^\circ$ by $52.8^\circ$, the $53^\circ$ cone in Fig.~\ref{fig:hardware})~\cite{amsosram_tmf8829},
and the openings of the PDMS fingertips (Fig.~\ref{fig:hardware}b) do not block any part of it.
Each reading collects the reflections of a train of laser pulses into a histogram of their arrival times.
The length of this train is set as a number of iterations, which we lowered from the default of 600{,}064 to 102{,}400.
Thus, a sensor's measurement is ready 14~ms after its trigger instead of 37~ms (medians).
Reading the 256 zone values out of the sensor and sending them to the computer add about 15~ms.
Because of this sequential triggering, each sensor delivered new readings at 8.6~Hz on the four-sensor board and 11.4~Hz on the
three-sensor board in the recorded demonstrations (medians over the 150 demonstrations).
A ROS node publishes each sensor's latest reading at 30~Hz, held between updates (no interpolation). At test time no actions are sent
while any sensor's latest reading is older than 2~s.
Because the ToF readings drift while the sensors warm up, the demonstrations and robot trials started only after the sensors had
run for several minutes.

The sensor reports each zone's distance in steps of 0.25~mm.
The sensor reports a zone only when its signal-to-noise ratio reaches the firmware's confidence threshold of 6.
The policy also marks a zone invalid when its calibrated distance is below 8~mm, for example, when a fingertip touches the Pringles can.
The computer corrects each zone with its own lookup table from measured to true distance, interpolated linearly.
We built the tables for each sensor with the preset and iteration count above.
A motorized linear stage placed a gray card at 20 distances from 10 to 400~mm, with 100 frames per distance.
At three distances left out of the fit (37, 75 and 225~mm), the tables reduce the mean absolute error of the median distance over
the central $8\times8$ zones from 19.7~mm to 0.35~mm (0.16 to 0.46~mm per sensor).
Over all 256 zones, the mean absolute error per zone falls from 30.2~mm to 0.52~mm.
Because these are errors of 100-frame medians, single readings vary more, with a frame-to-frame standard deviation of about 0.9~mm
per central zone from 15 to 75~mm and 2.7~mm at 400~mm.

\subsection{How Dark the Camera Images Are}
\label{app:dark}
\begin{figure*}[t]
\centering
\includegraphics[width=\textwidth]{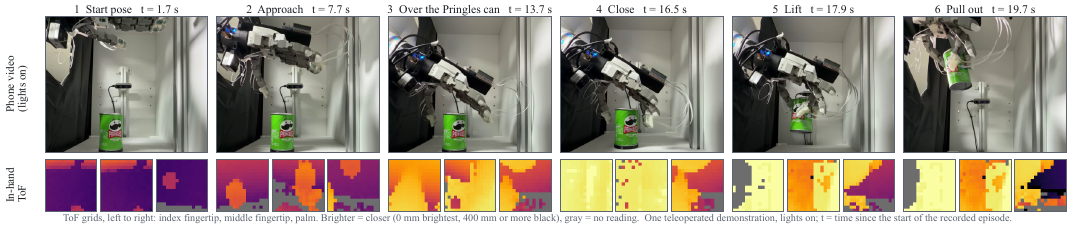}
\caption{\textbf{The task in one teleoperated demonstration, lit for illustration only} (all demonstrations and robot trials ran in the dark):
(1) start pose, (2) approach, (3) over the Pringles can, (4) close, (5) lift, (6) pull out. $t$ is the time since the episode start.
Bottom: the three ToF grids the policy reads (index fingertip, middle fingertip, palm). Brighter = closer, gray = no reading.}
\label{fig:timestrip}
\end{figure*}
The lux meter of Section~\ref{sec:demos} is a handheld WT81 digital meter, whose lowest range (0 to 199.9~lux) reads in steps of 0.1~lux.
Its stated accuracy below 10{,}000~lux is $\pm$3\,\% of the reading.
Its reading of 0.0~lux was taken with the meter and the hand in the shelf, and with the LEDs of the hand and the ToF boards uncovered.
The egocentric camera, outside the curtain, shows the arm's indicator lights faintly but not the Pringles can.
In the video-compressed palm images the policy receives, the brightest 1\,\% of pixels have values of at most 2 on the 0 to 255
grayscale, so these images are black.
The wrist images are as dark in most of the first 54 demonstrations, but in 14 of them the brightest 1\,\% of pixels reach values of up to 185.
These bright pixels show the blue curtain cloth rather than the Pringles can.
Only the uncompressed frames, never given to the policy, show the Pringles can very faintly.

\subsection{Teleoperation in the Dark}
\label{app:teleop}
Fig.~\ref{fig:timestrip} shows the task in one teleoperated demonstration, lit for illustration only.

\begin{figure}[t]
\centering
\includegraphics[width=\columnwidth]{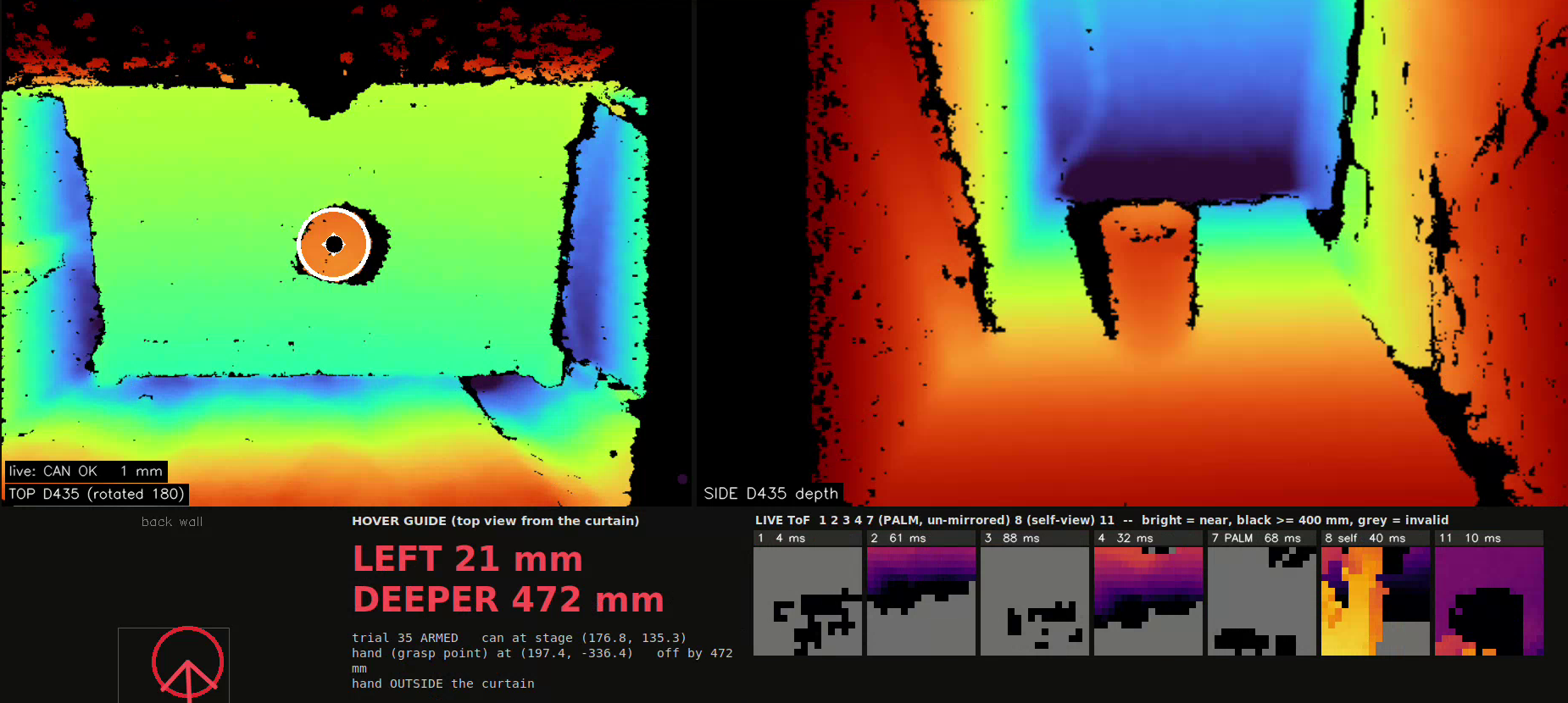}
\caption{\textbf{The operator's screen during teleoperation in the dark.} Left, depth top view from the shelf-top camera, with the
Pringles can circled. Right, depth side view. Bottom, the drawn guide with the remaining distance in $x$ and $y$, and the live
ToF grids (brighter = closer, gray = no reading). Recorded before a demonstration, with the hand still outside the curtain.}
\label{fig:teleop_ui}
\end{figure}

The wrist tracker is located by Vive base stations, which emit their own infrared light.
The base stations stood outside the curtained shelf, aimed away from its opening, and were on in every demonstration and robot trial.
The operator's screen (Fig.~\ref{fig:teleop_ui}) shows two depth views and a drawn guide.
A RealSense D435i (a D435 with a built-in inertial sensor) in the shelf top gives the top view, and a D435 gives a side view of the hand's height above the Pringles can.
Both cameras compute depth by triangulation between their two built-in infrared sensors.
The infrared projectors of both cameras, which add a dot pattern to aid this triangulation, were on during the demonstrations and the robot trials.
The drawn top-view guide shows the Pringles can (from the $x$--$y$ stage), the hand (from the arm joint angles) and the
remaining distance in $x$ and $y$.

\textbf{Infrared check.} With the hand held still at the start pose, switching the projectors of both shelf cameras on and off changed the
typical zone's median distance by at most 0.6~mm on any sensor (0.2~mm on the palm sensor and 0.2~mm with nothing switched).
No zone changed by more than three times its frame-to-frame noise (one 15~s recording per condition, one hand pose, base stations on).

\textbf{Unused demonstrations.} Of the 60 demonstrations of the first round, the first 3 were recorded before the 2~s hold at the start pose
was added to every episode. Three more had recording faults. In one, all seven ToF sensors read about half the true distance, and in another, an
overloaded computer delayed the arm controller's messages by up to 83~ms. In the third, the palm ToF sensor froze. All six were
excluded before any policy was trained or tested.

\subsection{Training Settings}
\label{app:training}
All weights except the frozen ToF encoder are fine-tuned in full from the $\pi_{0.5}$ base model with a batch size of 32, at 3.27~s
per step on an RTX PRO 6000 GPU (2.91~s for the camera-only VLA).
The 54- and 94-demonstration policies are trained for 10 epochs (14{,}780 and 24{,}350 steps).
The 144-demonstration policies keep the 94-demonstration step count, about 7 epochs.
This step count was fixed before training because of time constraints.
Training uses a cosine learning-rate schedule with 1{,}000 warm-up steps, from $2.5\times10^{-5}$ to $2.5\times10^{-6}$ for the
pre-trained weights and from $10^{-4}$ to $10^{-5}$ for the new ones.
It keeps an exponential moving average of the weights with a decay of 0.999.
The ToF distances are scaled by a random factor drawn from $U(0.95,1.05)$, and the ToF input is never dropped.
Each demonstration begins with a 2~s hold at the start pose.
For the 54-demonstration policies, training begins each demonstration 0.5~s before the hand starts to move, so it keeps at most
the last 0.5~s of the start-pose hold (1.5~s for the policy with the joint angles in the prompt and the frozen encoder, whose 10
epochs therefore take 15{,}290 steps).

The ToF encoder is pre-trained for 3{,}000 steps to estimate the Pringles can's position from ToF readings taken with the hand at
the start pose.
The pre-training data are 3,166 readings at 201 positions, nearly all on a 15~mm grid, and 57 readings at 56 positions scanned in the dark. The encoder's output layer is then removed.
The ToF expert is a randomly initialized transformer of width 512 and depth 18 (99.1\,M parameters).

The 40 additional demonstrations of the second round raise the number of demonstrations within 3~cm of the two front-corner positions of the
8-position pilot robot tests from 2 and 4 to 10 and 11.
Here, front means the curtain side, nearest the robot, and these two positions are not among the 16 test positions of
Fig.~\ref{fig:positions}.

\subsection{Robot Trial Protocol}
\label{app:protocol}
\begin{figure}[t]
\centering
\includegraphics[width=\columnwidth]{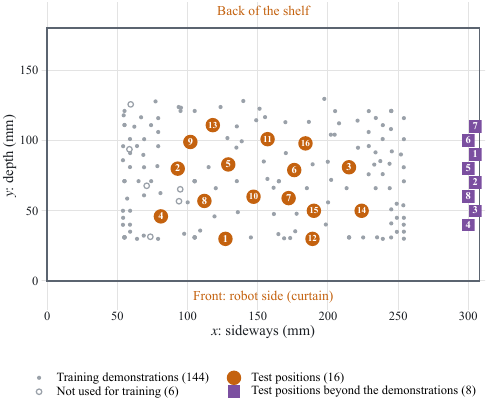}
\caption{\textbf{Pringles can positions, top view} (robot at the bottom, $x$--$y$ stage coordinates). Orange marks the 16 test positions, numbered in run
order. Gray dots mark the 144 training demonstrations and gray circles the 6 recorded demonstrations not used for training (Appendix~\ref{app:teleop}).
Purple squares mark the 8 positions beyond the demonstrated area (Section~\ref{sec:results}), numbered in run order.}
\label{fig:positions}
\end{figure}

\textbf{Trial plan.} Before the first night, the positions, the trial order, the approach rule computed from the joint logs and two tests
were written down and time-stamped. These tests compare the VPLA policy with the invalid-ToF control and with the camera-only VLA,
each by an exact McNemar test with Holm correction. A third test, the baseline VPLA policy against the VPLA policy, was fixed in
the same way before the baseline trials, after the first night's results were known. Every trial is reported.
Fig.~\ref{fig:positions} shows the 16 test positions together with the positions of the demonstrations.
It also shows the 8 positions beyond the demonstrated area, where the VPLA policy and the two PLA policies were tested on a later
night (Section~\ref{sec:results}).
The trials of the PLA policies at these 8 positions were added during that night, after the VPLA policy had approached at none of them.
They were still written down and time-stamped before they ran.

\textbf{Code names.} A script assigns each policy a random code name and hides the assignment. The operator sees no per-policy log, timing or
memory use until all trials are scored, although a policy's movements can still reveal which policy it is.

\textbf{Execution.} The robot-side client program sends the policy the three images,
compressed like the demonstration video, and the raw ToF readings.
The client then executes the first 8 actions of each 16-action chunk at 30~Hz. Joint steps
are limited to 0.025~rad (arm) and 0.12~rad (hand). Arm poses are relative to the last commanded arm pose.

\textbf{Safety limits} (the same for every policy).
(1) \emph{Contact stop.} If, for 3 control steps while the hand is open, the motor current of the index, middle, ring or little finger
is at least 60~mA and at least 20~mA above that finger's level 0.5 to 2~s earlier, the arm is not allowed to go lower. The hand and the policy
continue, and the limit is lifted when the hand closes. If the hand is still open 12~s later, the arm backs up along its last 0.5~s of
motion and the policy replans. The contact stop never fired in any of the 70 trials of the first two nights (48 blinded, 16 of the baseline VPLA policy and
6 of the PLA policies).
(2) \emph{Shelf-top limit.} The wrist stays at most 10~mm above the highest wrist position of the first 54 demonstrations at the
same or a greater $y$.
(3) \emph{Workspace box.} The hand stays within the space the demonstrations used, plus a 15~mm margin.
Targets that break (2) or (3) are not sent, and the policy replans. After 10~s of refused targets in a row, the arm and hand hold
still, and the operator ends the trial and scores it.

\textbf{Scoring.} The operator scores each trial from 0 to 4. A score of 0 means a collision or no approach, 1 a failed approach,
2 a successful approach but a failed grasp, 3 a successful grasp (the Pringles can held after the hand first closes) but a failed lift
or pull-out, and 4 the Pringles can taken out. Grasp means a score of at least 3 and pull-out a score of 4. The approach is computed
from the joint logs (Section~\ref{sec:setup}).

\textbf{Stopping rules.} The operator stops a trial only for safety, and not before the hand has come down 50~mm unless a collision is
imminent. A trial stopped for safety is scored by the stage it reached before the stop. A failed first grasp ends the trial (no second
attempt). Only a trial with an operator error or an equipment fault, or one in which the policy received input for less than 2~s, is
discarded and rerun, at most twice per policy.

\textbf{Blinding.} Code names could not hide how each policy moved. The invalid-ToF control and the camera-only VLA came down near
the left wall in most trials, as the operator had seen in an earlier session.
In contrast, the VPLA policy came down over the Pringles can in every trial. Thus, the operator could tell the VPLA policy from the controls, and the grasp and pull-out scores may not be fully blind.
The approach, computed from the joint logs, is not affected.

\textbf{Baseline night.} The baseline VPLA policy ran on the following night, without code names, at the same 16 Pringles can
positions and in the same order as the VPLA policy.
Because its trials ran on another night, we checked afterwards that both policies started from the same inputs.
The arm pose was the same in every trial, and the hand joint angles differed by at most 1.8$^\circ$.
At the same Pringles can position, the start-pose ToF readings of the two nights differed by a median of 4.7~mm, less than the median
difference of 7.0~mm between readings at different Pringles can positions.

\textbf{PLA policies (descriptive).} The PLA policies receive neither camera images nor joint angles.
On the first night, after the 48 trials, each ran 3 unblinded trials at the first three test positions.
Together they approached in all 6 trials and grasped in 5, but pulled the Pringles can out in only 2.
In 2 of the 3 failed lifts, the hand kept rising toward the shelf top until the workspace box refused further targets, so we
stopped testing both policies for safety.
A likely reason for these failed lifts is that all three ToF sensors face the Pringles can once the hand grasps it, so the input of a PLA policy barely
changes during the lift.
On a later night, a safety limit refused every commanded pose that would raise a point of the arm or hand more than 10~mm above
its height at the start pose.
This limit was meant mainly for the lift, so that the hand could not collide with the shelf top.
It also ruled out the pull-out, so the operator ended each trial once the hand held the Pringles can clearly off the shelf floor.
Both PLA policies then ran unblinded at all 16 test positions (Section~\ref{sec:results}).
Their approaches and grasps exceeded the 3 approaches and 0 grasps of the camera-only VLA, with $p \le 0.0005$ in each of the four
exact McNemar tests after Holm correction.

\subsection{Exchange Test Details}
\label{app:exchange}
Section~\ref{sec:offline} describes the exchange test.
The motion of a demonstration starts at the first of three frames in a row in which the grasp point moves horizontally faster than 20~mm/s.
The same 9 demonstrations supply the exchanged ToF inputs at every test frame.
Their Pringles cans stood nearest the 9 grid points with $x$ = 73, 160 or 237~mm and $y$ = 37, 91 or 119~mm, which cover the area
of the tested demonstrations.
A tested demonstration that is itself one of the 9 has only 8 exchanges, since an exchange with its own ToF input would change nothing.
The displacement $d_i$ comes from the $x$--$y$ stage positions of the two Pringles cans.
For the human demonstrations, the slope of the recorded grasp-point motion over the next 16 actions against the Pringles can
position, at the same three test frames, is 11.1\,\% in $x$ and 14.9\,\% in $y$.
Unlike the ToF sensitivity, this slope compares different demonstrations, each with its own recorded motion and Pringles can position.
The baseline VPLA policy trained on 54 demonstrations also gives a ToF sensitivity of 0.0 at 5{,}000 and 10{,}000 steps, so this result
does not depend on the training length.
All exchange-test numbers are open loop, on training frames, for one training seed.

\begin{table}[t]\footnotesize
\centering
\caption{\textbf{Full ablation.} ToF sensitivity in percent as in Table~\ref{tab:ablation}, with 95\,\% intervals below the table. The encoder is trained
from scratch with the policy, pre-trained and then trained further (not frozen), or pre-trained and frozen. Robot: number of the 16 test positions approached / grasped / pulled out, from three nights,
blinded unless marked. n/a: no ToF input. --: not tested.}
\label{tab:ablation_full}
\setlength{\tabcolsep}{3.5pt}
\begin{tabularx}{\columnwidth}{@{}c c c c X@{}}
\toprule
Joint angles & Encoder & Demos & ToF sensitivity & Robot \\
\midrule
in prompt & from scratch & 54 & 0.0 / 0.0 & -- \\
in prompt & frozen & 54 & 0.0 / 0.0$^a$ & -- \\
none & from scratch & 54 & 1.5 / 1.3 & -- \\
none & frozen & 54 & 3.3 / 3.7 & 9/16, 2/16, 2/16 \\
none & frozen & 94 & 6.7 / 8.4 & 13/16, 13/16, 11/16 \\
none & frozen & 144 & \textbf{8.7 / 9.0} & \textbf{16/16, 16/16, 16/16} \\
in prompt & from scratch & 144 & 0.0 / 0.0 & 0/16, 0/16, 0/16$^b$ \\
in prompt & frozen & 144 & 0.1 / 0.2 & -- \\
none & from scratch & 144 & 6.5 / 5.1 & 7/16, 7/16, 7/16 \\
none & pre-trained$^c$ & 144 & 7.6 / 9.0 & -- \\
\midrule
\multicolumn{3}{@{}l}{Invalid-ToF control (144)} & 0.0 & 1/16, 0/16, 0/16 \\
\multicolumn{3}{@{}l}{PLA policy, images masked (94)} & 11.8 / 12.6 & 16/16, 12/16, --$^d$ \\
\multicolumn{3}{@{}l}{PLA policy, images masked (144)} & 11.5 / 11.5 & 16/16, 14/16, --$^d$ \\
\multicolumn{3}{@{}l}{Camera-only VLA (144)} & n/a & 3/16, 0/16, 0/16 \\
\multicolumn{3}{@{}l}{Human demonstrations} & 11.1 / 14.9 & -- \\
\bottomrule
\end{tabularx}
\\[2pt]
\raggedright\scriptsize
95\,\% intervals without joint angles: 54 demonstrations 1.2 to 1.8 / 0.9 to 1.8 (from scratch) and 2.5 to 4.3 / 2.6 to 4.8 (frozen), 94
demonstrations 5.8 to 7.7 / 7.0 to 9.8, 144 demonstrations 5.5 to 7.5 / 4.1 to 6.2 (from scratch), 6.8 to 8.5 / 7.7 to 10.4 (pre-trained) and
7.7 to 9.7 / 7.5 to 10.6 (frozen). With
joint angles, 144 demonstrations and the frozen encoder, 0.08 to 0.13 / 0.15 to 0.25, and all other rows with joint angles within
$\pm$0.01.
$^a$Demonstrations start 1~s earlier (15{,}290 training steps). $^b$Following night, unblinded. $^c$Pre-trained, not frozen
(offline only). $^d$Unblinded, under a safety limit on the hand's height that ruled out the pull-out (Appendix~\ref{app:protocol}).
\end{table}

\end{document}